# Distributed JEPA: A Self-Supervised Framework for Energy Forecasting

**Liana Toderean[1], Tudor Cioara[1]*, Vasilis Michalakopoulos[2], Efstathios Sarantinopoulos[2], Ionut Anghel[1], Elissaios Sarmas[2]**

[1]Distributed Systems Research Laboratory, Computer Science Department, Technical University of Cluj-Napoca, G. Barițiu 26-28, 400027 Cluj-Napoca, Romania; liana.toderean@cs.utcluj.ro, tudor.cioara@cs.utcluj.ro, ionut.anghel@cs.utcluj.ro.

[2]Decision Support Systems Laboratory, School of Electrical & Computer Engineering, National Technical University of Athens, Ir. Politechniou 9, 157 73 Athens, Greece; vmichalakopoulos@epu.ntua.gr, ssarantinopoulos@epu.ntua.gr, esarmas@epu.ntua.gr.

*Corresponding author: tudor.cioara@cs.utcluj.ro

**Abstract:** Traditional energy forecasting solutions rely on task-specific supervision and energy asset representations, limiting transferability and the ability to capture general temporal dynamics across heterogeneous assets. We address this by proposing a distributed Joint Embedding Predictive Architecture (JEPA) for self-supervised learning from heterogeneous energy time-series. The framework predicts latent representations of masked temporal segments while integrating temporal observations and contextual information within a shared embedding space. To prevent representation collapse, training combines a latent-space predictive objective with covariance and temporal variance regularization. The evaluation was conducted on energy consumption and generation datasets under data-degradation scenarios and compared with a Transformer forecasting baseline. The learned representations remained stable (cosine similarity ≈0.98; effective rank 185–235). JEPA achieved performance comparable to a Transformer on building energy data, higher $R^2$ in 3/5 consumer clusters, and outperformed the baseline on 9/10 unseen PVs ($R^2$=0.73–0.88 vs. <0.45), while showing greater robustness to missing data.

## 1. Introduction

Traditionally, forecasting problems in the energy domain have been tackled using specialized models designed for an individual site or energy asset. However, this specialization can limit the transfer of knowledge between related forecasting problems and fail to take advantage of common structural patterns in energy time-series [1]. As a result, there is growing interest in unified time-series representation learners that can extract shared temporal dynamics and support multiple downstream objectives within a single framework [2, 3]. However, privacy and regulatory constraints often limit access to fine-grained measurements, while differences in asset types, geographical locations, and operating conditions introduce domain changes among datasets. In parallel, recent advances in transformer-based architectures have substantially improved forecasting accuracy by introducing more expressive sequence modeling capabilities. Existing approaches include efficient attention variants [4], decomposition-based models [5], patch-based tokenisation [6], variate-wise attention [7], and temporal 2D representations [8], with evidence that simpler decomposed linear models can also be highly competitive [9] and emphasising the importance of appropriate inductive biases over model complexity alone. However, existing models are trained in a fully supervised manner for a specific forecasting task and optimize objectives that minimize prediction errors at individual time steps [10]. As a result, they often require large amounts of labeled data and have limited transferability across energy domains [11]. Additionally, autoregressive or reconstruction-based objectives encourage models to focus on low-level signal reconstruction which may favor low-level statistical features over higher-level temporal dynamics that generalize across settings [12].

These limitations suggest that further progress may depend less on increasingly specialized forecasting architectures and more on learning general-purpose representations of time-series. Self-supervised learning facilitates this by leveraging large amounts of unlabeled data to learn transferable representations [13]. Franceschi et al. [14] introduce a triplet-loss framework based on temporal proximity, while Temporal Neighborhood Coding (TNC) [15] formalises this via distinguishing nearby from distant temporal windows. CoST [16] extends contrastive learning with season–trend disentanglement, yielding representations suited to structured signals such as building energy data. TS2Vec [17] employs hierarchical contrastive objectives at multiple granularities to produce general-purpose contextual embeddings, and TF-C [18] improves cross-domain transfer by aligning time-domain and frequency-domain representations for stronger few-shot generalisation. Although several federated learning approaches have been proposed for supervised forecasting models to address privacy concerns [19], existing self-supervised representation learning methods are typically designed for centralized training on pooled datasets. As result, they cannot be easily deployed in distributed energy environments, where privacy constraints and statistical heterogeneity prevent data sharing across entities.

Among emerging representation-learning paradigms, Joint Embedding Predictive Architectures (JEPA) differ from reconstruction-based [20] and contrastive representation learning approaches [21] by employing a predictive objective in latent space. Rather than directly reconstructing future observations, JEPA learns to predict latent representations, encouraging the model to focus on invariant temporal structure and high-level dynamics [22]. This self-supervised formulation has the potential to produce transferable embeddings that generalize across energy assets and forecasting tasks while reducing dependence on large, labeled datasets [23]. Despite its promise, only a limited number of state-of-the-art methods currently implement the JEPA paradigm, with most existing approaches focusing on image and video modeling. I-JEPA [24] and its video extensions [25, 26] demonstrate that latent-space prediction can learn rich semantic and temporal representations without reconstruction. Similar success has been reported in audio [27, 28], language [29, 30], and user-interface modeling [31], highlighting the versatility of predictive representation learning across modalities and motivating its application to time-series data. For time-series domain, Ennadir et al. propose an adapted JEPA [12] for temporal data by masking non-overlapping patches and predicting the latent representations of missing segments, evaluating the learned features on classification and forecasting tasks. He et al. [32] extend this idea with a multi-resolution architecture and a soft codebook bottleneck to capture long-term trends and regularize the latent space. However, both approaches remain vulnerable to embedding collapse, where representations either become nearly identical or occupy only a low-dimensional subspace. Although existing methods address this through heuristics such as stop-gradient [33], Balestriero and LeCun [34] showed that an isotropic Gaussian distribution is optimal for minimizing downstream prediction risk for JEPA embeddings, and by introducing LeJEPA, an objective that prevents collapse by constraining the embeddings towards this distribution. Since EMA-based training can suffer from representation collapse [35], Mo and Tong propose combining JEPA with contrastive learning to encourage more discriminative embeddings.

To address these limitations, we propose a distributed JEPA for energy time-series representation learning with hierarchical decoders for downstream prediction tasks (**Figure 1**). Instead of optimizing a task-specific forecasting objective, the model learns to predict latent representations of masked temporal segments, decoupling representation learning from downstream tasks while encoding temporal energy dynamics and static contextual information in a unified embedding space that supports multi-modal integration. This self-supervised formulation reduces dependence on annotated datasets and enables adaptation to new forecasting tasks. Reconstruction of raw observations is handled by downstream decoders, preventing the model from fitting high-frequency noise and asset-specific fluctuations, and encouraging general representations that capture structured temporal dynamics, long-range dependencies, and cross-modal relationships across heterogeneous energy assets and operating conditions. To ensure representation diversity and prevent collapse, the model is trained with a cosine similarity objective in latent space, alongside covariance and temporal variance regularization, which reduce redundancy across batch instances and time steps. The distributed design enables local encoding at the asset level, avoiding centralized raw data collection and making the approach suitable for privacy-sensitive settings. We analyze training dynamics and evaluate 24-hour-ahead forecasting accuracy across varying data-degradation scenarios, comparing against a Transformer baseline trained specifically for the same task under identical settings. The model performs comparably on a building energy consumption dataset, while achieving higher performance on the photovoltaic dataset with fewer available samples and a smaller decrease in forecasting accuracy across most degradation settings.

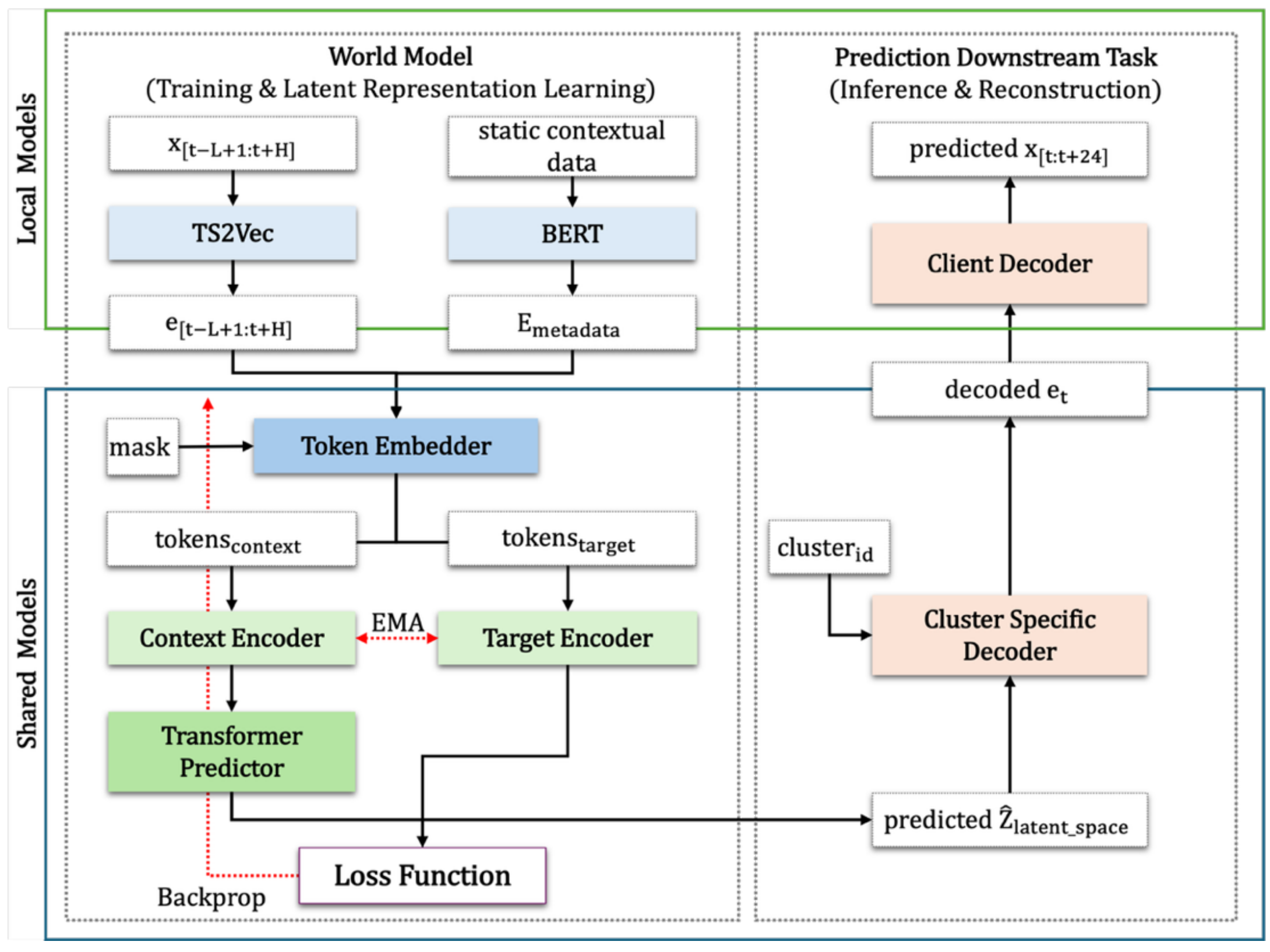

**Figure 1 Overview of the distributed energy time-series JEPA architecture**

## 2. Methods

The proposed model adapts the JEPA architecture to learn a unified representation of the energy domain by integrating multiple data modalities. Specifically, it combines energy time-series measurements with static contextual information (e.g. building characteristics, PV site specification). To transform inputs into a common embedding space with fixed dimensionality, modality-specific encoders are used. A temporal encoder (TS2Vec) processes time-series consumption data, while a contextual encoder (BERT) transforms static metadata into semantic embeddings. The architecture is designed to be extensible, so that additional data modalities and their corresponding encoders can be incorporated when additional information is available. The Token Embedder creates the tokens using a mask for the target positions, providing the tokens for the Context Encoder (contextual data embedding and context window time-series embeddings) and for the Target Encoder (target time-series embeddings). During training, the output of the Transformer Predictor is compared to the representations produced by the Target Encoder. The resulting loss is used to update the parameters of the Token Embedder, Context Encoder and Transformer Predictor to enforce learned representations of the observed context that encapsulate relevant temporal and contextual information. Finally, the parameters of the Target Encoder are updated as an exponential moving average (EMA) of the Context Encoder parameters.

### 2.1. JEPA Model for energy data

The training process for the JEPA transformer encoders and predictor is illustrated in **Figure 2a**. Firstly, the input time-series and static metadata embeddings are organized into batches that align multiple clients within the same temporal window. Each sample consists of a historical context window of length L and a target prediction horizon of length H starting at time step t.

The Token Embedder generates two distinct sequences by processing the raw input embeddings through separate logic paths. For the Context Encoder, the system uses the Future Mask to select only the visible indices (metadata and the L context embeddings). Only this subset is passed through the projection layer and combined with positional embeddings, ensuring that the target is entirely excluded from the context representation. For the Target Encoder, the Token Embedder processes the full, unmasked time-series and metadata embeddings. In this path, the entire sequence is projected and augmented with positional embeddings to serve as input for Target Encoder. The Context Encoder is a 6-layer standard transformer that produces contextualized representations of visible positions.

A predictor network then cross-attends to the latent representations of the context and produces predictions at the H masked positions. Since masked positions are absent from the context encoder output, the predictor uses learned positional embeddings as query vectors to represent the target positions during cross-attention.

The full token sequence is passed through Target Encoder, an exponential moving average (EMA) copy of the Context Encoder that receives no gradient updates directly. It produces latent representation for the entire sequence and the ones from target positions are used as the ground-truth latent representations for the Loss Function.

The Loss Function has a composite objective designed to minimize predictive error while preventing representation collapse. The primary loss $\mathrm{L_{cos}}$ measures the cosine similarity between the $\mathrm{L_2}$-normalized latent representation generated by the predictor $\hat{\mathrm{Z}}$ and the latent targets $\mathrm{Z}$, generated by the Target Encoder:

$$\mathrm{L_{cos}} = 1 - \frac{1}{\mathrm{B \cdot H}} \sum_{\mathrm{i=1}}^{\mathrm{B \cdot H}} \frac{\hat{\mathrm{Z}}_\mathrm{i} \cdot \mathrm{Z_i}}{\left\|\hat{\mathrm{Z}}_\mathrm{i}\right\|_2 \left\|\mathrm{Z_i}\right\|_2} \tag{1}$$

To ensure the learned representations are informative and will not collapse during training, two regularization terms are applied. First, a covariance regularization term $\mathrm{L_{cov}}$ penalizes the off-diagonal elements of the covariance matrix C of the flattened predictions (across batch and time) to maximize feature diversity:

$$L_{cov} = \frac{1}{D}\sum_{i\neq j}[C]_{i,j}^2 \tag{2}$$

Second, a temporal variance term $L_{tvar}$ ensures the model captures dynamic changes across the H sequence steps. It uses a hinge loss to maintain the variance σ of the predictions above a threshold ε:

$$L_{tvar} = \frac{1}{B\cdot D}\sum_{b=1}^{B}\sum_{j=1}^{D}\max(0, \varepsilon - \mathrm{Var}_h(\hat{z}_{b,j})) \tag{3}$$

where $\mathrm{Var}_h$ is the variance calculated over the prediction time dimension.

Third, $L_{norm}$ penalises the mismatch in magnitude between each predicted target embedding and its corresponding EMA target, averaged over all predicted timesteps in the batch:

$$L_{norm} = \frac{1}{B\cdot H}\sum_{i=1}^{B\cdot H}(\|\hat{z}_i\|_2 - \|z_i\|_2)^2 \tag{4}$$

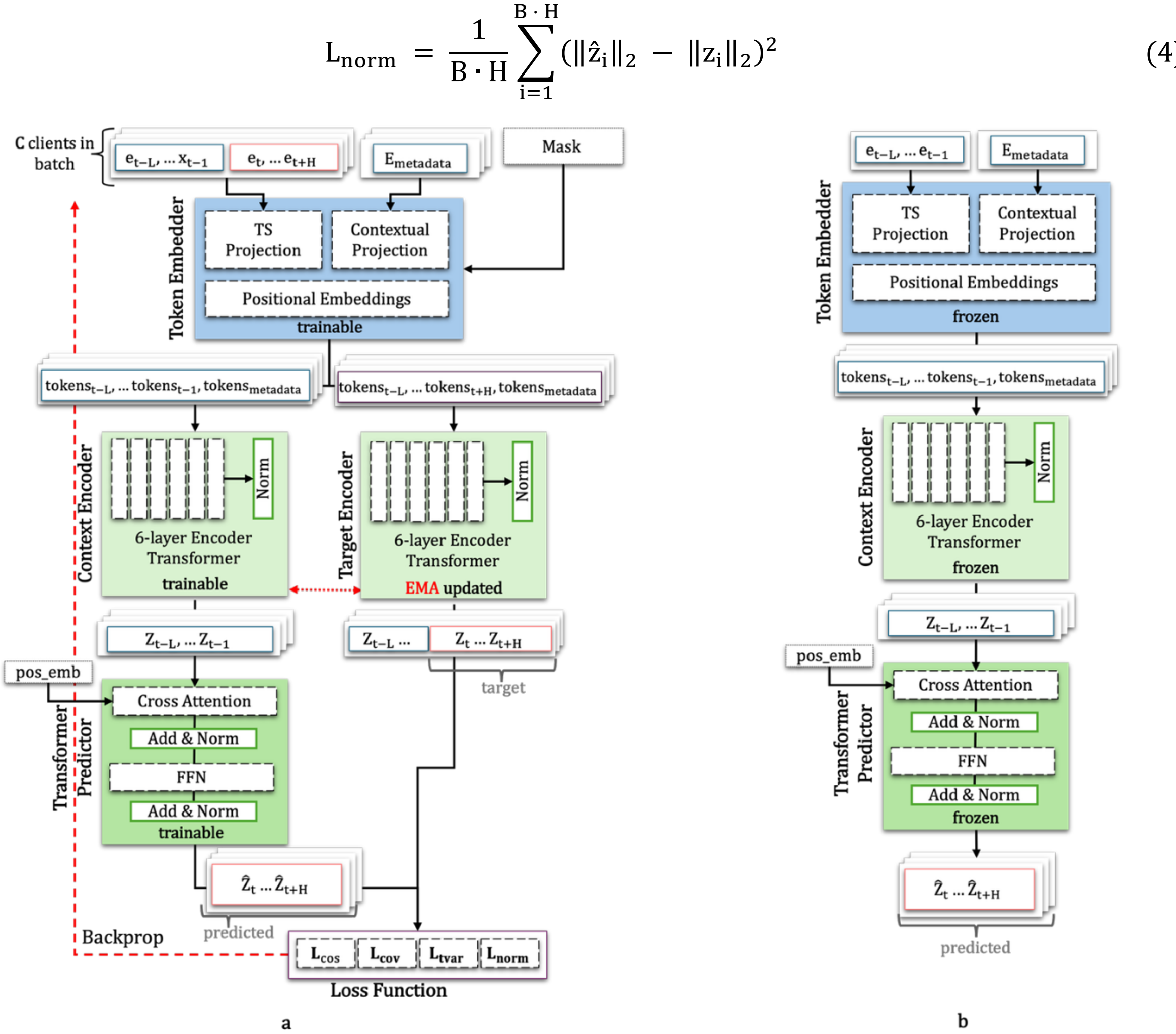

**Figure 2 JEPA World Model detailed architecture (a) for training the JEPA Encoders and Predictor (b) for embedding generation using the frozen Context Encoder and Predictor**

The training process follows a decoupled update strategy. During the forward pass, gradients are tracked only for the Token Embedder, Context Encoder, and Predictor. The backpropagation updates these components using AdamW optimizer using the composed JEPA loss:

$$\mathrm{Loss_{JEPA}} = \mathrm{L_{cos}} + \alpha \mathrm{L_{cov}} + \beta \mathrm{L_{tvar}} + \gamma \mathrm{L_{norm}} \tag{5}$$

The Target Encoder parameters $\theta_{target}$ are updated once per batch using the EMA of the context encoder's *parameters* $\theta_{context}$:

$$\theta_{target} \leftarrow m\theta_{target} + (1 - m)\theta_{context} \tag{6}$$

After the training of the JEPA transformers, the Target Encoder is discarded as its role was to generate the targets for the Loss Function. The inference pipeline of the generation phase is presented in **Figure 2b**. The input embeddings $e_{t-L} \dots e_{t-1}$ are passed through the Token Embedder and Context Encoder to obtain the latent representation $z_{t-L} \dots z_{t-1}$ of the visible context. The latent representation can be used as features for various downstream tasks, such as forecasting, classification, or anomaly detection. In the specific case of forecasting, the Transformer Predictor is kept as it learned during training to predict future latent embeddings $\hat{Z}$ for the target horizon.

## 1.1. Forecasting downstream task

To complete the downstream forecasting task, the predicted latent embeddings generated by the Predictor Transformer must be decoded into raw time-series values. Since the time-series data is distributed across multiple clients, a hierarchical decoder structure is used, as illustrated in **Figure 3**, allowing clients to use a local decoder model.

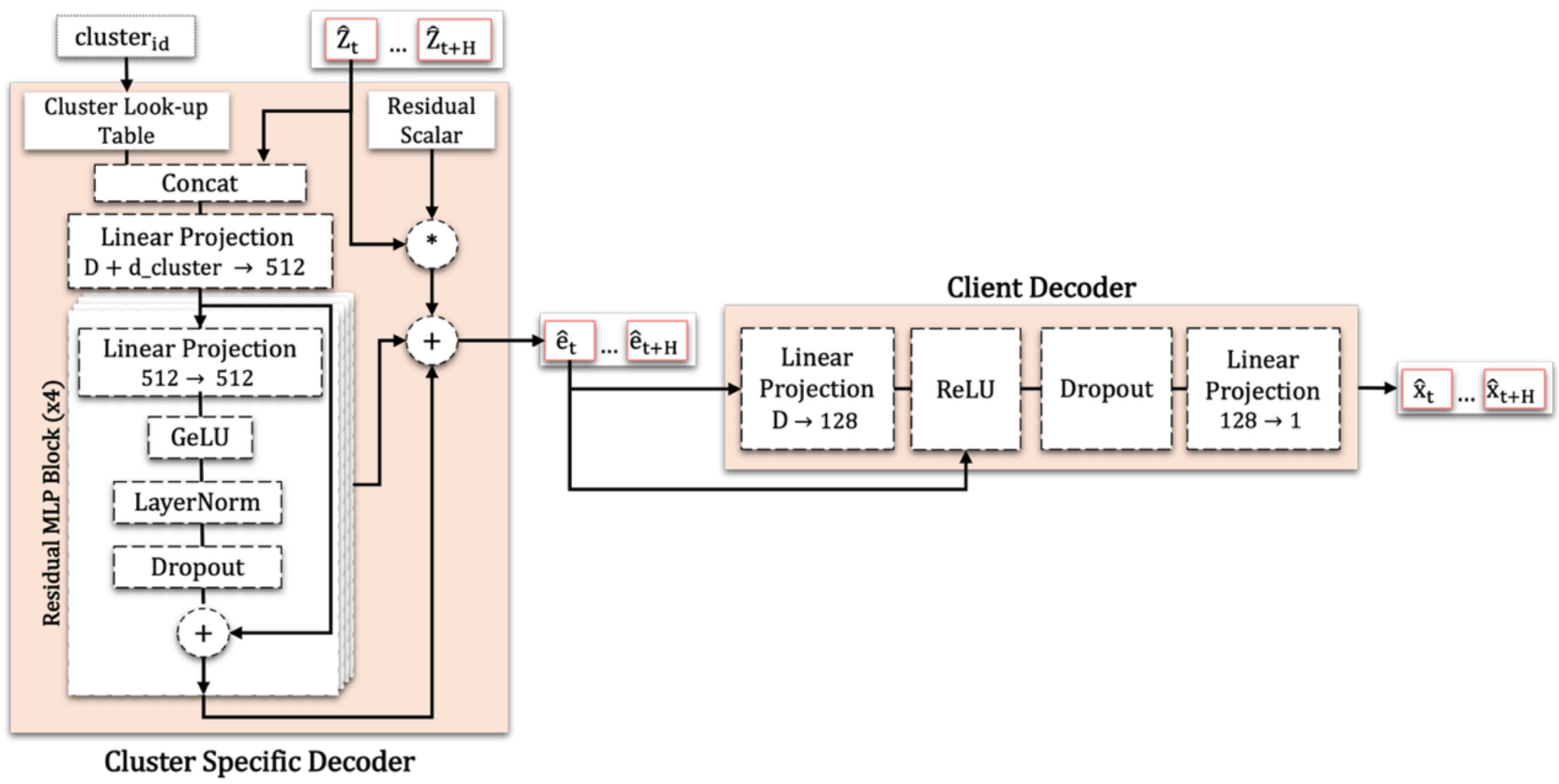

**Figure 3 Prediction Downstream Task Decoders**

First, a Cluster Specific Decoder maps the predicted latent embeddings into the TS2Vec embedding space. The clustering applied to TS2Vec embeddings separates different time-series patterns or behaviors, allowing the decoder to learn distinct transformation rules conditioned on the cluster id. Each client applies its own local decoder to convert the decoded TS2Vec embeddings into raw time-series values. During training, the cluster decoder learns a general mapping from latent representations to TS2Vec embeddings, while the client decoders learn the final reconstruction based on their local data. The Cluster Specific Decoder retrieves a learned vector from a Cluster Look-up Table based on the cluster_id, which is then concatenated with the predicted latent embeddings $\hat{Z}$. This representation passes through a Linear Projection and a stack of four Residual MLP Blocks, each containing GeLU activation, LayerNorm, and Dropout. Then the output of the MLP blocks is added to the original input embedding, which have been scaled by a Residual Scalar, to produce the decoded TS2Vec embeddings ê. The Client Decoder has a simple structure consisting of a Linear Projection layer, ReLU activation, Dropout, and a final projection layer. It has as input the decoded TS2Vec embedding ê and outputs the raw time-series values $\hat{x}$.

To evaluate prediction quality in the latent space, the decoders were trained to reconstruct a target from a single embedding rather than the entire prediction horizon. This formulation isolates reconstruction performance from temporal dynamics and enables assessment of the information encoded in the latent representation. The decoders were trained using JEPA latent embeddings and the corresponding TS2Vec embeddings computed on the training set. During inference, the decoder was applied independently to each predicted timestep to reconstruct the full forecasting horizon.

### 1.2. Local TS2Vec embedding generation and clustering

TS2Vec is a self-supervised encoder for time-series representation learning, trained through contrastive learning. To ensure both consistent time-series representations and the privacy of sensitive consumer energy data, TS2Vec is trained sequentially across all clients. On every client, the model is firstly initialized with the shared weights and then optimized independently using only the client's local data. After local training, the updated model weights are shared, while the raw energy data is kept private at the client. This approach enables learning a unified representation space across clients while preserving data privacy.

The energy time-series dataset is denoted as $[x_0, \dots, x_T]$ where $T$ is the total number of time steps in the series. The dataset is divided into train and test sets considering the predefined parameters train_size and test_size. Raw energy time-series values of the client are sampled in windows of size $L$, for which TS2Vec generates two overlapping subseries (views) by cropping different segments of the original window. The TS2Vec encoder then produces two sequences of embeddings for each view $Z'$ and $Z''$, with each embedding vector being of dimension $D$. The contrastive loss ensures that the model learns to generate similar embeddings for the overlapping portions of the views. After the TS2Vec model is trained its weights are shared with all the clients. Then, each client samples windows from its time-series denoted as $W_t = [x_{t-L}, \dots, x_t]$, where t is the target position. The windows are passed through the local TS2Vec model to generate representation embeddings for each time step. The model outputs a sequence of embeddings $E_{seq} = [e_{t-L}, \dots, e_t]$, and the representation of the energy at timestep t is the embedding array $e_t$. The process for training and generating TS2Vec embeddings on energy time-series is presented in **Figure 4**.

After the TS2Vec representations (embeddings) are generated, they are used to cluster the households' prosumers based on their distinct energy profile features, based on previous research efforts [36, 37, 38, 39]. To segment the data based on the features of each load profile, three distinct clustering algorithms are evaluated: K-means, K-medoids, and Hierarchical clustering. Because the optimal number of clusters in such analyses cannot typically be known in advance, these algorithms are systematically tested over a predefined range of candidate clusters, from k=2 to k=30. This extensive range is explored to determine the most appropriate structural configuration using three established evaluation metrics: the Silhouette Score (SIL), the Davies-Bouldin Index (DBI), and the Calinski-Harabasz Index (CHI). By assessing the algorithms across these metrics, the optimal clustering algorithm and the ideal number of distinct prosumer segments can be objectively identified for final application.

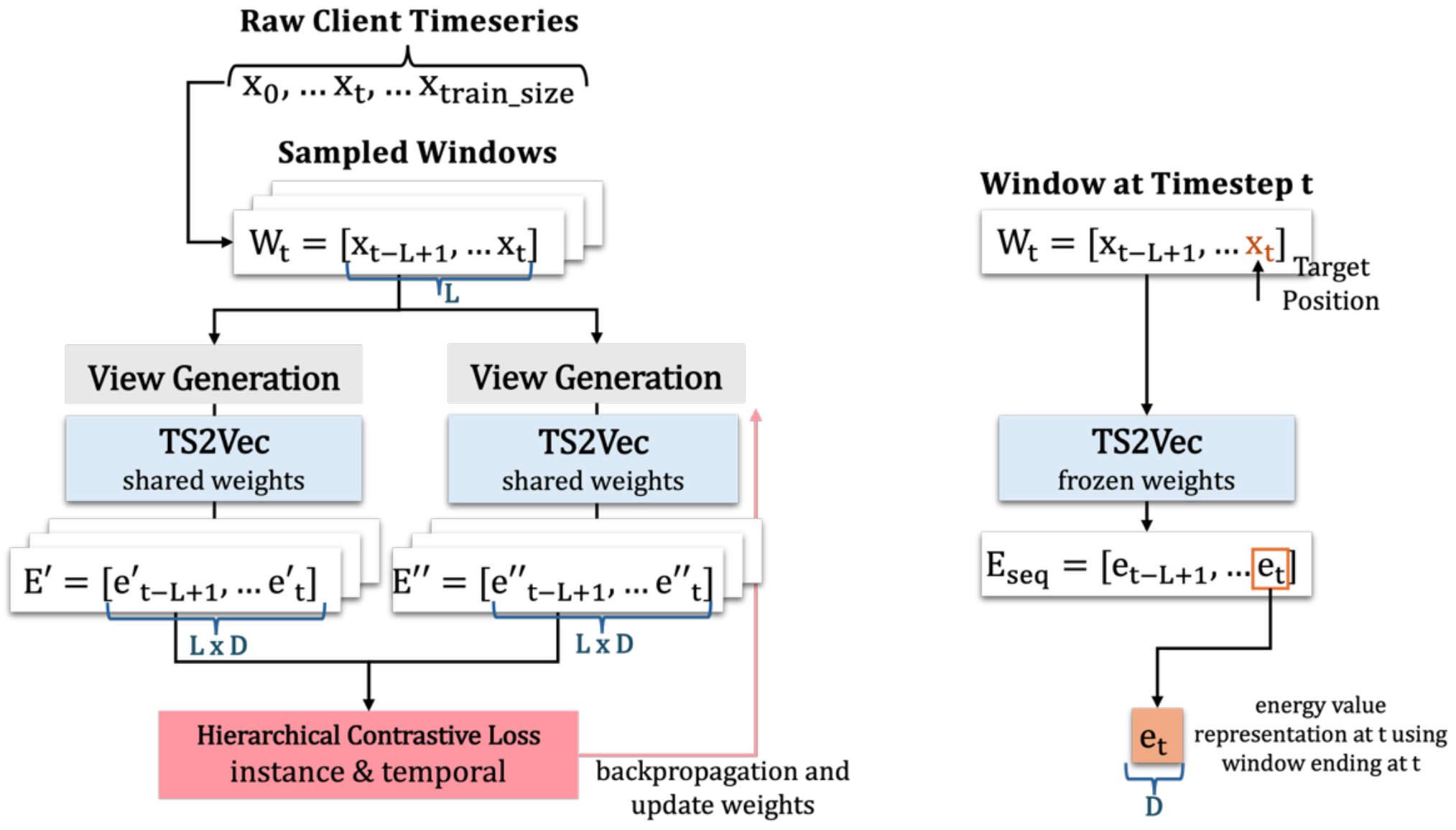

**Figure 4 TS2Vec time series encoder training (left) and generation (right)**

## 1.3. Local BERT Embedding Generation

The pre-trained BERT model is used to generate contextual embeddings from metadata capturing the semantic relationships between different data types (**Figure 5**). The metadata is processed from CSV files containing multiple columns that indicate the information available through two primary data types: text and numeric. While text columns are extracted directly, numeric columns undergo a discretization process to transform continuous values into discrete tokens, ensuring compatibility with the BERT vocabulary. The processed features are then serialized into a JSON string and passed through the BERT Tokenizer. The resulting tokens are processed by the pre-trained BERT model with frozen weights that extracts by default a raw embedding $Z^{\beta e}{}_{rt}$ of size 768. Finally, a trainable projection layer maps this vector into a D-dimensional space, designed to match the dimensions of the time-series embeddings.

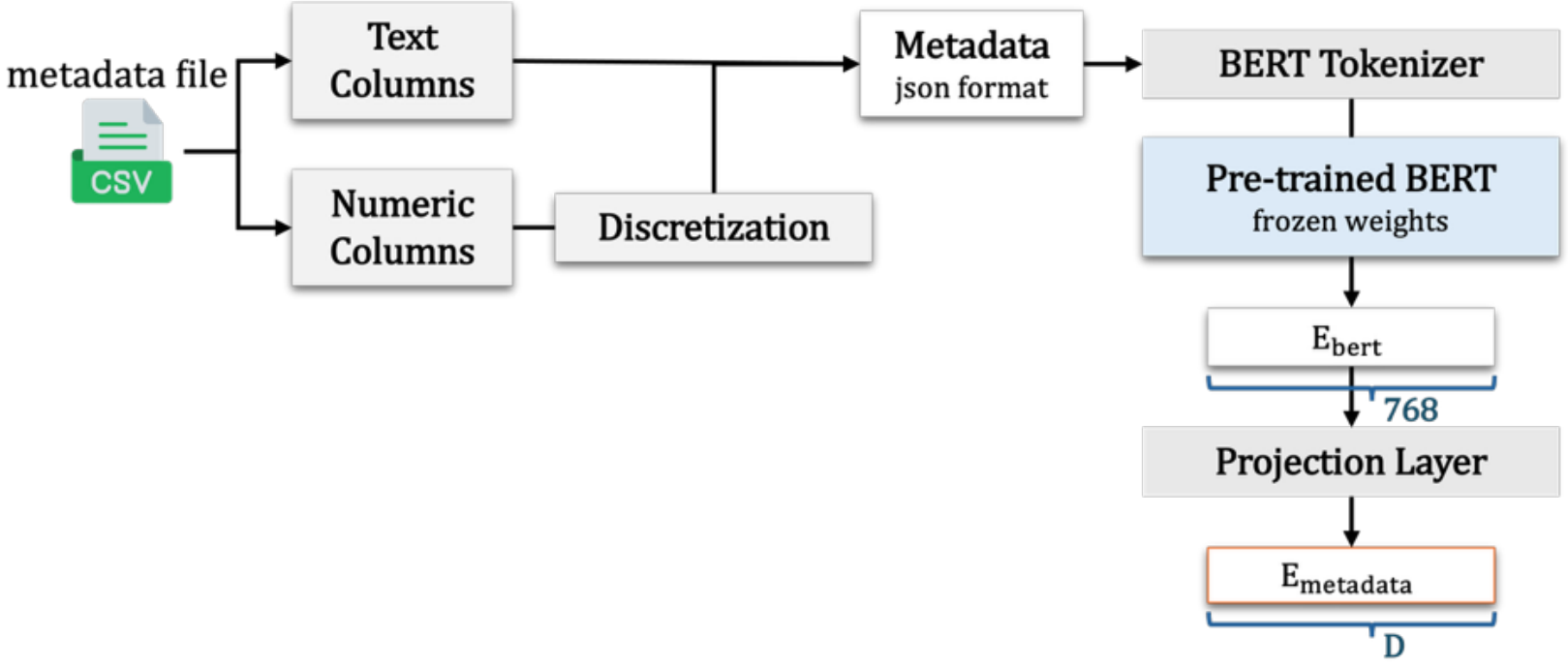

**Figure 5 BERT contextual embedding generation**

## 1.4. Evaluation Methodology

For the evaluation of JEPA latent representations, we considered the 24h-ahead forecasting downstream task. The goal is to assess the performance of the JEPA learned representations on the forecasting task compared to a Transformer baseline method trained on TS2Vec representations under identical data conditions.

Both datasets are split following the same procedure. Firstly, the clients are split into 80% training clients and 20% test clients, where the test clients are never seen during any stage of JEPA training. For the training clients, the available data is further divided into 80% training, 10% validation, and 10% test splits. The training set is used for training the JEPA models and for training the cluster decoder. For the test clients, a separate 70/15/15 split is applied per client, where the 70% portion is used to fine-tune the cluster-specific decoder and to train the individual TS2Vec decoder, while the remaining 30% is used for validation and testing of the adaptation performance.

The proposed JEPA approach first learns latent representations through a Transformer-based predictor operating in embedding space. The latent representations are then decoded into raw energy values using a separate decoding module, enabling a structured mapping from learned latent dynamics to the forecasting target. The JEPA architecture and decoder training hyperparameters are summarized in **Table 1** including the input/output specifications, architectural configuration of each component, and decoder training settings shared between the Genome and PV datasets ("train" and "ft" denote pretraining and fine-tuning stages, respectively).

**Table 1 JEPA model architecture and shared hyperparameters**

| Category | Hyperparameter | Value |
|---|---|---|
| **Input/Output** | Context length L | 168 |
| | Forecast horizon H | 24 |
| **Architecture** | *Token Embedder* | |
| | Input/Output/$d_{dim}$ | 256/256/256 |
| | Max sequence length | 512 |
| | *Context, Target Enc. & Predictor* | |
| | $\mathrm{d_{model}}$ | 256 |
| | Attention heads h | 8 |
| | Encoder layers | 6 |
| | Predictor layers | 2 |
| | Feedforward dim $\mathrm{d_{ff}}$ | 1024 |
| | *Cluster-Specific Decoder* | |
| | MLP residual blocks | 4 |
| | Hidden dimension | 512 |
| | *Client Decoder* | |
| | Linear layers | 2 |
| | Hidden dimension | 128 |
| **Decoder Training** | *Cluster-Specific Decoder* | |
| | Batch size | 256 |
| | Learning rate | $1 * 10^{-3}$ (train) $1 * 10^{-4}$ (ft) |
| | Optimizer | Adam |
| | Epoch | 50(train), 20(ft) |
| | Loss | $0.5 * \cos + \mathrm{MSE}$ |

| | *Client Decoder* | |
|---|---|---|
| | Batch size | 256 |
| | Learning rate | $1 * 10^{-3}$ |
| | Optimizer | Adam |
| | Epoch | 50 |
| | Loss | MSE |

The training configurations specific to each dataset are listed in **Table 2**, highlighting the differences between the Genome and PV experimental setups and S0, S1, S2 correspond to the JEPA training stages for the PV dataset.

**Table 2 JEPA configuration for Genome and PV**

| **Category** | **Hyperparameter** | **Genome** | **PV** | | |
|---|---|---|---|---|---|
| | | | ***S0*** | ***S1*** | ***S2*** |
| **JEPA Training** | Dropout | 0.1 | 0.1 | | |
| | Masking strategy | Future | Var. blocks | Future | Future |
| | EMA Momentum | 0.999 | 0.996 → 0.999 | | |
| | Learning rate | $3 * 10^{-4}$ | $3 * 10^{-4}$ | $1 * 10^{-4}$ | $5 * 10^{-5}$ |
| | Batch size | 1024 | 512 | | |
| | Epochs | 50 | 20 | 10 | 10 |
| | Loss weight α | 0.04 | 0.10 | 0.05 | 0 |
| | Loss weight β | 1.0 | 1 | 0 | 0 |
| | Loss weight γ | 0 | 0 | 0.5 | 0 |

As a baseline model, we trained an encoder-only version of a standard Transformer [40]. The model has TS2Vec representations as input and outputs a direct 24h energy forecast using a Transformer encoder with sinusoidal positional encodings followed by a lightweight prediction head. To ensure a strong reference model, we conducted a hyperparameter search over several architectural and training parameters, including the model dimension, number of attention heads, feed-forward dimension, dropout rate, learning rate, and batch size. The configuration achieving the lowest validation MAE was selected for evaluation, and the resulting hyperparameter configuration is summarized in **Table 3**. The datasets were split following the same procedure previously described. To ensure a fair comparison with the JEPA-based approach, the Transformer baseline was trained using the same TS2Vec representations in two phases. First, a global model was trained using data from all training clients. Then, the encoder was frozen and only the prediction head was fine-tuned on the first 70% of each test client's data, allowing the model to adapt to client-specific generation patterns before evaluation.

**Table 3 Baseline Transformer model hyperparameter**

| **Category** | **Hyperparameter** | **Value** |
|---|---|---|
| **Input/Output** | Embedding dimension D | 256 |
| | Context length L | 168 |
| | Forecast horizon H | 24 |
| **Architecture** | Model dimension $d_{model}$ | 128 |
| | Attention heads | 8 |
| | Encoder layers | 2 |
| | Feedforward dim $d_{ff}$ | 256 |
| | Activation function | GELU |
| | Positional encoding | Sinusiodal |
| | Prediction head | LN→ Linear (128,128) → GELU → Dropout → Linear (128,24) |
| **Training** | Batch size | 128 |
| | Epochs | 50 |
| | Learning rate | $5 * 10^{-4}$ |
| | Optimizer | AdamW |
| | Loss | MSE |
| | Patience | 5 |
| **Head fine-tuning** | Learning rate | $5 * 10^{-4}$ |
| | Epochs | 20 |
| | Patience | 3 |

In addition to the standard evaluation performed on clean data, we evaluated the forecasting performance under three data degradation scenarios applied to the 256-dimensional TS2Vec embedding windows. The first scenario is additive Gaussian noise, where independent noise from a normal distribution with mean zero and standard deviation relative to each client distribution is added to every element of the embedding vectors. The second scenario simulates intermittent sensor dropout through random missing data, where each timestep in a window is independently set to zero with a probability between 0.0 and 0.50. The third scenario simulates longer sensor outages by removing contiguous blocks from the input. A continuous segment of timesteps is removed from each window simulating longer sensor outages. The length of the missing block is varied, and its starting position is randomly selected within the window.

## 2. Results

We evaluated the proposed JEPA architecture on two publicly available energy datasets considering 24 hours ahead energy prediction task: the Building Data Genome Project 2 dataset [41] and a rooftop photovoltaic (PV) dataset [42], as described in the Methods section.

### 2.1. Self-supervised learning dynamics

We examined the learning dynamics of the JEPA architecture on both datasets to evaluate the latent representation quality, avoid representation collapse, and ensure convergence during distributed training. The losses and similarity metrics are not sufficient to evaluate the representation quality, as in the context of JEPA it can also signal a partial collapse where the target and context encoders converge toward redundant, low-information representations, making the prediction task easy. Thus, we used the effective rank [43, 44] to track if the model uses the full dimensionality of the embedding vector and the uniformity metric [45] to measure how uniformly the learned features are distributed on the unit hypersphere. The metric evolution during training for both datasets is represented in **Figure 6a,b**.

The training and validation loss for the Genome dataset monitored across epochs show a simultaneous decrease in both loss curves with a lower validation loss caused by the absence of training noise (dropout and batch shuffling) together with the fact that the target encoder is updated only during training, leading to more consistent and predictable representations during evaluation. The cosine similarity reached ≈0.98 indicating strong alignment between the predicted and target representations. The effective rank stabilizes between 185–200 out of 256 dimensions, indicating that the model utilizes approximately 75% of the available embedding capacity. The uniformity metric [45] slightly increases from −3.5 to −2.9 during training. As can be seen, by epoch 15 the per-epoch improvement in validation loss had reduced to less than 0.002, justifying the stopping criterion.

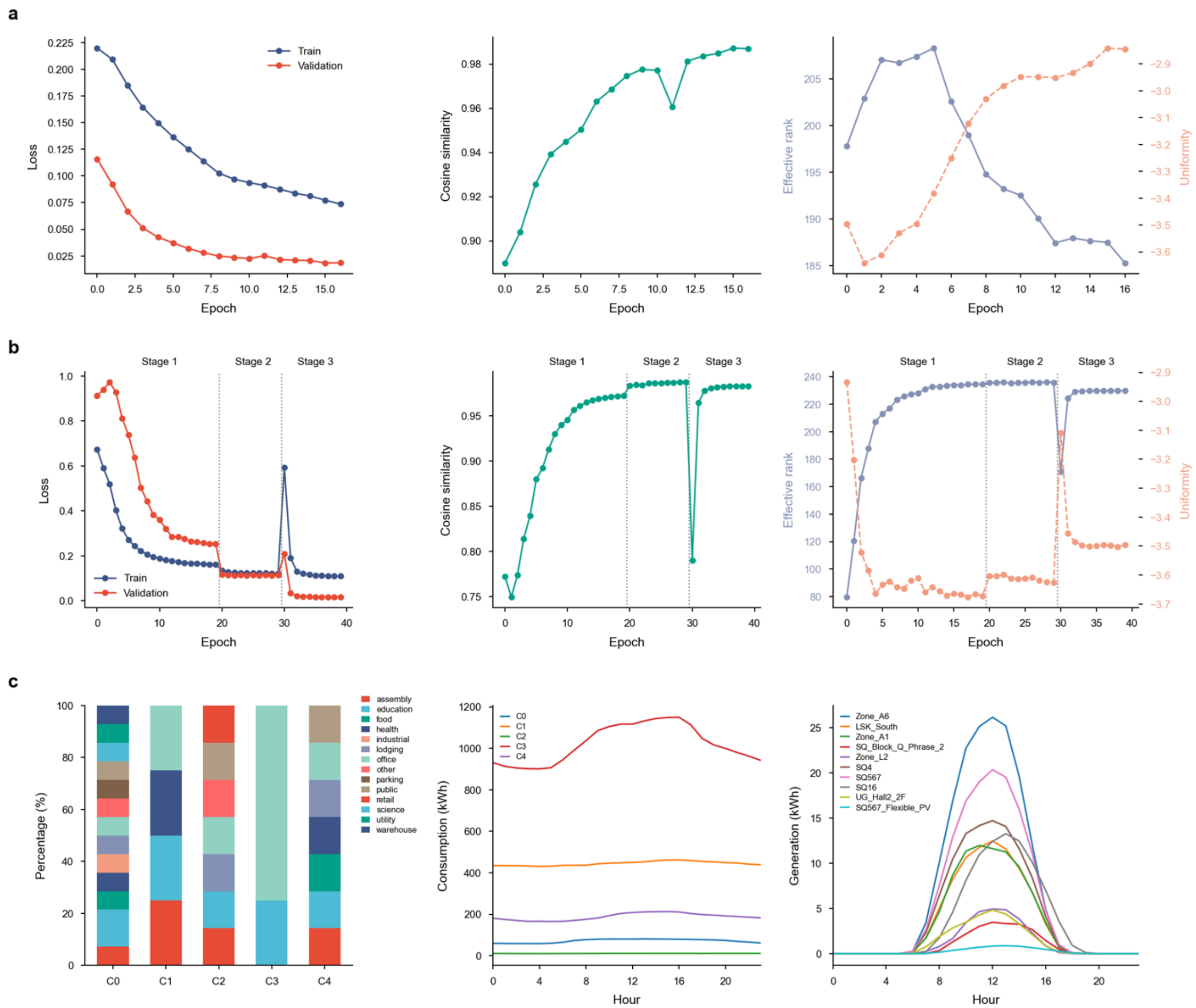

**Figure 6 JEPA training dynamics and test set characterization for Genome and PV (a)** Genome (Future Mask) and **(b)** PV (Stage 1: Random Multi-block Masking Stage 2: Future Mask + norm loss increase Stage 3: Future Mask - encoders frozen, train new predictor). **(c)** Genome building type distribution across cluster (left), mean hourly consumption per cluster (middle), and mean hourly PV generation per test client (right).

Considering the lower diversity of the PV dataset, the training was divided into three stages, in contrast to the Genome dataset, where training was performed in a single stage using only future masking. In the first stage, all models were trained using random multi-block masking, enabling the encoder to learn general latent representations without being tied to a specific forecasting objective. During this stage, the validation loss decreased from 0.91 to 0.25, cosine similarity increased to 0.97, and the effective rank increased to 235. In the second stage, training was continued using only future masking, while increasing the weight of the norm loss to preserve the amplitude of the embeddings. The validation loss and cosine similarity remained stable, while effective rank and uniformity showed small changes, indicating that the representation geometry was preserved while the model adapted to future representation prediction. In the third stage, the encoders were frozen and a newly initialized predictor was trained using future masking. This resulted in an initial decrease in cosine similarity and effective rank due to the cold start of the predictor, followed by rapid recovery within a single epoch and convergence to the lowest loss achieved across all three stages. These results demonstrate that the general representations learned during the first stage can be effectively used for future representation prediction.

### 2.2. Transfer learning capabilities

To evaluate the transfer capabilities of the JEPA latent embeddings, we selected a subset of 36 consumers from the Genome test set (N = 208) and used all 10 stations from the PV test set. Due to the limited number of stations available in the PV dataset, clustering was not performed, and the complete test set was used for evaluation. For the Genome dataset, the test consumers were selected by sampling a fixed percentage of consumers from each cluster, such that the relative distribution of clusters in the subset matches that of the full test set, while enforcing a minimum number of consumers for smaller clusters to ensure fair representation. In addition, this sampling strategy preserves the building-type diversity within clusters, as illustrated in **Figure 6c** (left), alongside the mean energy values for each cluster (middle) and the mean generation profiles for the PV (right).

To compare prediction performance on the train set against the test consumers and PV sites that were not used for JEPA training, we selected a subset of 44 consumers from the Genome train set to reflect a comparable cluster composition to the test set and used all 40 training PV sites. Across all five clusters, JEPA achieves a median $R^2$ of 0.648–0.854 on training consumers and 0.660–0.844 on test consumers, with differences in median $R^2$ within 0.02–0.137 across clusters. Test performance is comparable or slightly exceeds training performance in four of the five clusters (C0: 0.840 vs 0.854; C2: 0.718 vs 0.703; C3: 0.785 vs 0.648; C4: 0.844 vs 0.842). For the PV dataset, the median $R^2$ was 0.818 on test versus 0.838 on training, with a small drop in performance of 0.02. The $R^2$ median along with the std are represented in **Figure 7**. Together, these results indicate that JEPA learns transferable representations across heterogeneous consumers and sites.

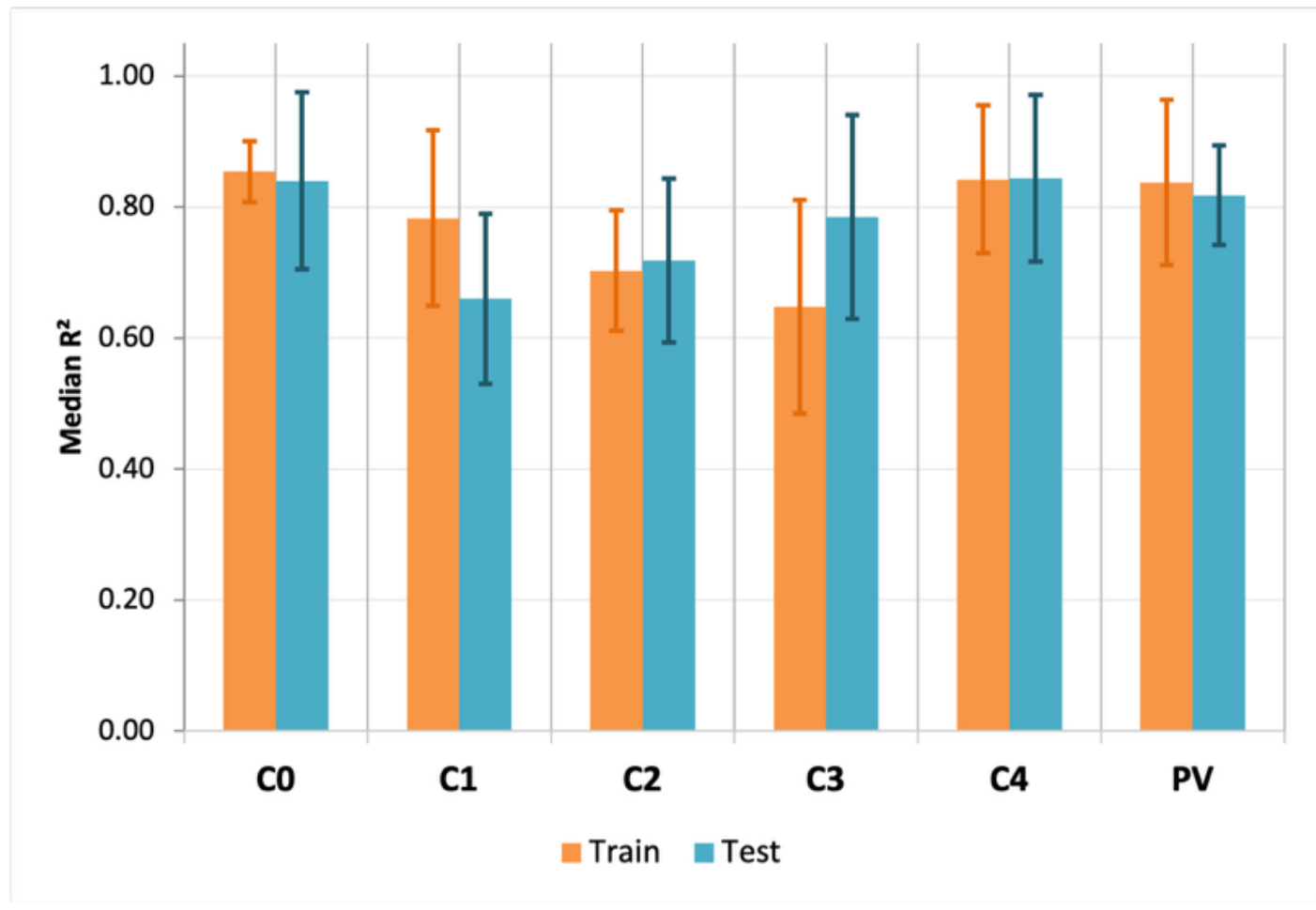

**Figure 7 JEPA transferability Evaluation on train vs test sets**

In **Figure 8** are represented the reconstructed energy values form the latent space prediction (Predicted) and the actual measured values (Actual) for representative buildings from each cluster and PV sites from the test set.

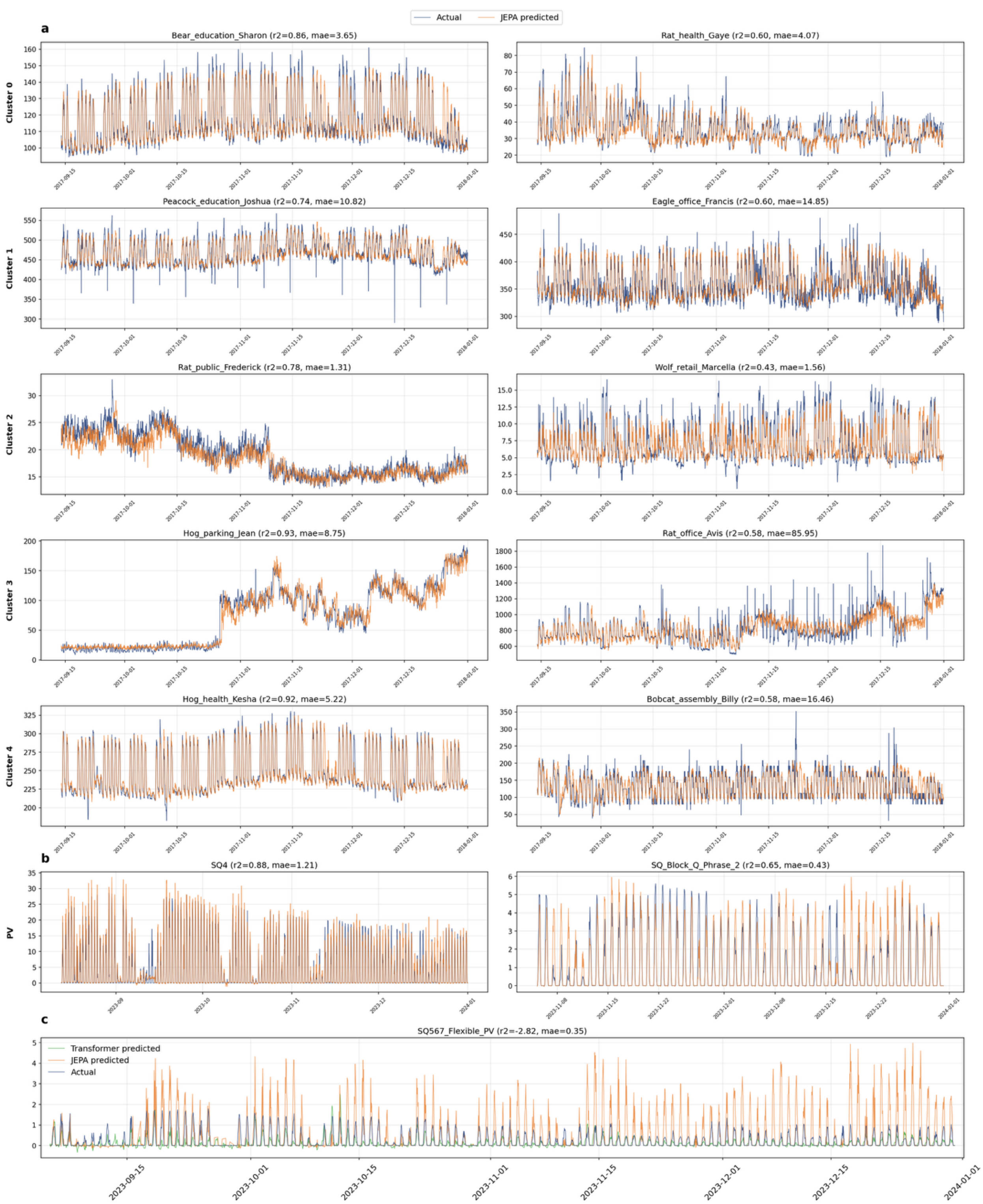

**Figure 8 Actual energy values and JEPA predicted values for representative Genome consumers and PV Sites: (a) Genome dataset (b) PV dataset (c) SQ567_Flexible_PV**

The buildings and sites shown on the left correspond to low MAE and high $R^2$, while the ones on the right have a higher MAE and relatively lower $R^2$. These plots, correlated with the evaluation metrics, indicate that the model in general follows the temporal dynamics of consumption, with some difficulties in capturing high fluctuations and variations.

## 2.3. Energy prediction accuracy

**Table 4** shows the accuracy evaluation metrics (MAE, RMSE, and $R^2$) for both datasets, using a 168h context window for both JEPA and the Transformer baseline. On the Genome dataset, the average error metrics are computed per cluster, and overall performance is similar between JEPA and the baseline. The Transformer achieves lower MAE in all five clusters, though generally the differences are small relative to the cluster magnitudes (the largest absolute gap is in Cluster3, where errors are an order of magnitude larger than in other clusters). For RMSE, JEPA is lower in Cluster0 and Cluster1, while the Transformer is lower in Clusters 2, 3, and 4. For $R^2$, JEPA outperforms the Transformer in 3 of 5 clusters (Cluster0, Cluster1, Cluster4).

The PV dataset has substantially fewer distinct sites than Genome (40 training PV sites vs. 828 training buildings) with slightly longer time-series length per site (≈19.2k time steps for each site and ≈17.5k hourly timesteps for each building) and the metrics are reported individually for each PV site. JEPA significantly outperforms the Transformer on 9 of 10 PV sites. For several sites, the $R^2$ of the Transformer is below 0.45, and for Zone_A6 close to zero (0.018) while JEPA achieves $R^2$ between 0.73 and 0.88 on the same sites. This suggests JEPA learns transferable time-series dynamics that generalize well across PV sites with comparable, continuous generation profiles.

**Table 4 JEPA vs Baseline Transformer prediction performance comparison on Genome and PV datasets**

| Dataset | | MAE JEPA | MAE Transf. | RMSE JEPA | RMSE Transf. | $R^2$ JEPA | $R^2$ Transf. |
|---|---|---|---|---|---|---|---|
| **Genome** | Cluster0 | 4.746 | **4.691** | **6.542** | 6.782 | **0.840** | 0.819 |
| | Cluster1 | 12.832 | **12.694** | **17.738** | 18.082 | **0.660** | 0.637 |
| | Cluster2 | 1.310 | **1.096** | 1.712 | **1.451** | 0.718 | **0.724** |
| | Cluster3 | 62.970 | **55.654** | 91.586 | **85.552** | 0.785 | **0.801** |
| | Cluster4 | 8.290 | **8.062** | 10.842 | **10.753** | **0.844** | 0.843 |
| **PV** | LSK_South | **0.971** | 1.662 | **2.141** | 3.281 | **0.803** | 0.433 |
| | SQ16 | **1.131** | 1.202 | **2.280** | 2.385 | **0.844** | 0.817 |
| | SQ4 | **1.215** | 3.183 | **2.331** | 5.239 | **0.880** | 0.393 |
| | SQ567 | **1.612** | 4.358 | **3.214** | 7.663 | **0.872** | 0.274 |
| | SQ567_Flex_PV | 0.351 | **0.163** | 0.745 | **0.302** | −2.815 | **0.276** |
| | SQ_Block_Q_P2 | **0.432** | 0.563 | **0.862** | 1.013 | **0.649** | 0.532 |
| | UG_Hall2_2F | **0.494** | 1.054 | **1.036** | 1.865 | **0.755** | 0.144 |
| | Zone_A1 | **1.275** | 2.791 | **2.788** | 4.826 | **0.729** | 0.186 |
| | Zone_A6 | **2.283** | 7.374 | **4.581** | 11.688 | **0.843** | 0.018 |
| | Zone_L2 | **0.374** | 1.665 | **0.836** | 2.550 | **0.833** | −0.504 |

The exception is the SQ567_Flexible_PV, where the Transformer outperforms JEPA on all three metrics and the $R^2$ for JEPA is negative (−2.82). This site has a generation profile with much lower amplitude than other sites (see **Figure 6c**) and JEPA predictions overestimate the variability of this almost flat signal (see **Figure 8c**-SQ567_Flexible_PV). This result suggests that the learned representations may not fully capture the characteristics of low-amplitude generation profiles, potentially due to their limited representation in the training data.

## 2.4. Performance analysis under data degradation

The evaluation analysis for the data degradation scenarios is reported using the median and interquartile range (IQR), represented as the solid line and shaded region, respectively, in **Figure 9**. The results for the Genome dataset are on left and for the PV dataset on right. For each dataset, and type of data degradation indicated on each row (Gaussian Noise, Random Missing and Block Missing) the median ± IQR of each reported error metric (MAE, RMSE and $R^2$) is plotted across the levels of degradation applied (std, missing rate, and block size).

For each degradation level, the median is computed across the client-level performance metrics, while the IQR, defined as the range between the 25th and 75th percentiles, quantifies the variability in performance across clients. For the Genome dataset, the Transformer is more resilient to additive Gaussian noise, maintaining roughly half the MAE of JEPA at the highest noise level and having a narrower IQR, indicating more consistent behavior across clients. In contrast, JEPA demonstrates higher resistance to random missing data, maintaining lower prediction errors and higher $R^2$ values than the Transformer as the corruption level increases, even in cases where the Transformer achieves comparable or slightly better performance on clean data. Both models are unaffected by contiguous block missing corruption, with only slight changes in errors and $R^2$ even for 24h gaps.

On the PV dataset, JEPA achieves a higher performance on clean data, while the Transformer has a high variability across clients, reflected in a wide IQR under all conditions. Across all data degradation scenarios, JEPA consistently maintains lower prediction errors and higher $R^2$ values than the Transformer. The steeper degradation for JEPA is a consequence of stronger performance on clean data, which provides a larger margin for performance decrease. However, JEPA remains the more accurate model throughout the entire corruption range. Similar to the Genome dataset, block-missing corruption has only a minor impact on both models.

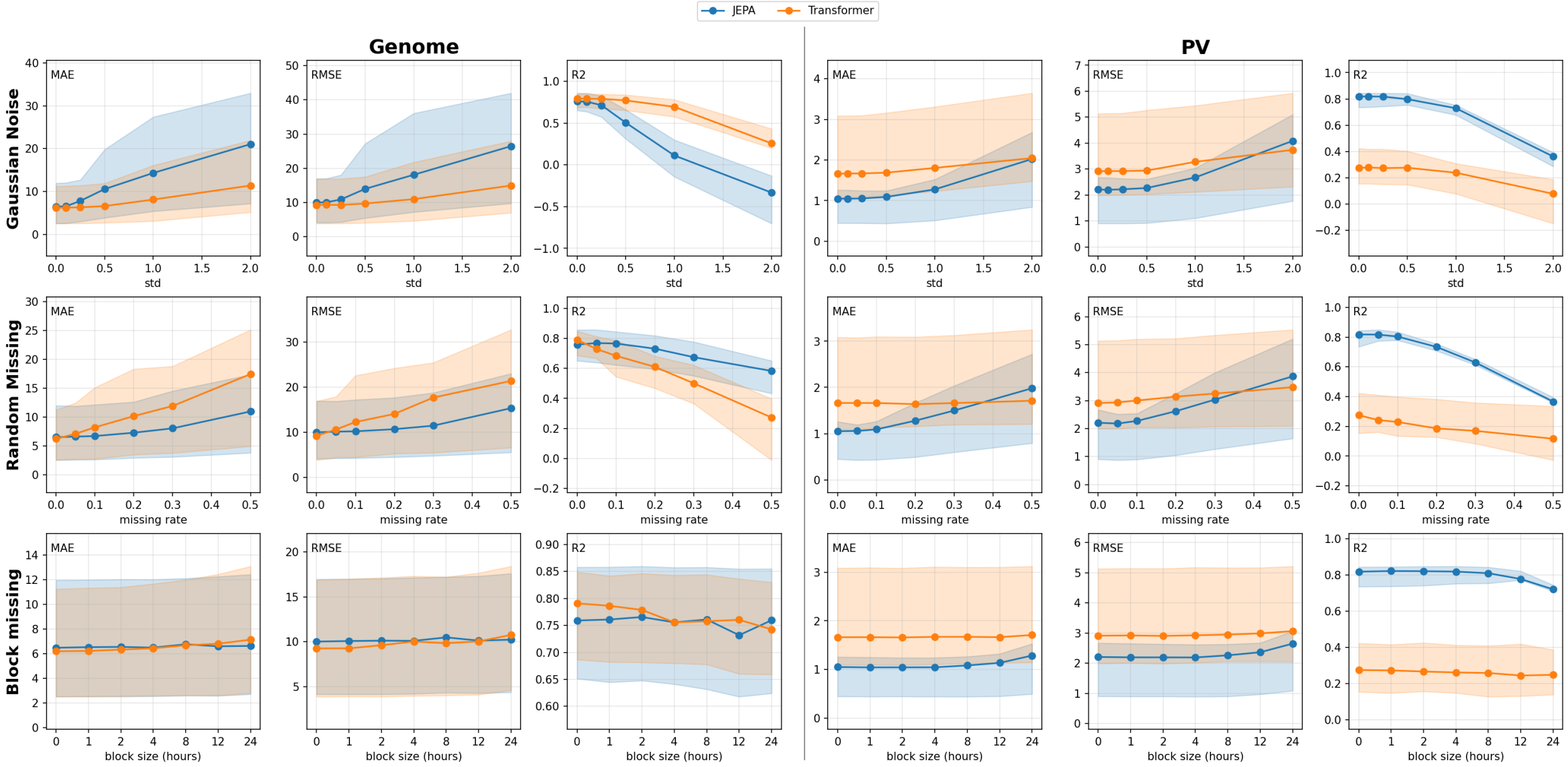

**Figure 9 Degradation of model performance under noise and missing data corruptions**

Overall, the results suggest that the Transformer is more tolerant to additive embedding space noise, particularly on the Genome dataset, which has a substantially larger training set. JEPA is more resilient when information is removed entirely, consistently maintaining lower errors and higher $R^2$ under random missing conditions on Genome, and retaining higher $R^2$ values on PV despite a steeper absolute decline.

## 3. Discussion

In this work, we proposed JEPA for self-supervised energy time-series representation learning, coupled with hierarchical decoders for downstream forecasting tasks. Rather than optimizing a task-specific forecasting objective, the model learns to predict latent representations of masked temporal segments, decoupling representation learning from downstream prediction while integrating temporal energy dynamics and static contextual information within a unified embedding space. We evaluated the proposed approach on building energy consumption and photovoltaic generation datasets under both clean and degraded data conditions.

The results indicate that self-supervised latent prediction constitutes a viable alternative to task-specific forecasting models and can produce stable and informative representations for heterogeneous energy time-series. Across both datasets, the proposed JEPA framework achieved high alignment between predicted and target embeddings while maintaining substantial latent-space diversity, indicating that the covariance and temporal variance regularization successfully mitigated representation collapse. The observed effective rank further suggests that the learned embeddings utilize a large fraction of the available representational capacity, supporting the hypothesis that latent-space prediction can capture meaningful temporal structure without relying on direct reconstruction of observations.

A notable finding is the transferability of the learned representations across distinct energy domains. The consistency of JEPA performance between train and test consumers and PV sites suggests that the model encodes transferable latent representations rather than consumer or site specific features. On the building energy consumption dataset, JEPA achieved performance comparable to a Transformer trained under identical conditions on TS2Vec representations, with both models exhibiting advantages on different client clusters. On the photovoltaic dataset, however, JEPA consistently outperformed the Transformer across most clients, achieving substantially lower forecasting errors and higher coefficients of determination. This result suggest that latent predictive objectives may be particularly beneficial in settings characterized by limited data availability, higher variability, or more complex temporal dependencies. These findings support the hypothesis that learning predictive representations in latent space encourages the extraction of more transferable temporal structures than directly optimizing forecasting objectives.

The energy forecasting accuracy and robustness analysis provides further insight into the properties of the learned representations. Although the Transformer had greater resilience to additive Gaussian noise on the building dataset, maintaining lower prediction errors at the highest noise levels, JEPA consistently outperformed the Transformer under random missing-data corruption. This behavior suggests that predicting latent representations of future temporal segments encourages the model to capture broader temporal dependencies and contextual information rather than relying heavily on individual observations. On the photovoltaic dataset, JEPA maintained lower errors and higher $R^2$ values across all degradation settings despite having a relative performance decline as corruption increased. Importantly, this degradation originated from a stronger clean-data baseline, and JEPA remained the more accurate model throughout the evaluated corruption range. Both models were largely unaffected by contiguous block-missing corruption, indicating that the TS2Vec representations preserve sufficient temporal context to tolerate moderate sensor outages.

The results also reveal several limitations. The proposed approach struggled on the SQ567 Flexible PV site, where generation patterns show substantially lower variability than those observed during training. The resulting performance degradation suggests that the learned representations may be less effective for rare or underrepresented operating regimes. In addition, the evaluation was restricted to building energy consumption and photovoltaic generation datasets and therefore does not fully characterize the generalization capabilities of the framework across the broader range of energy forecasting applications.

Future work should investigate larger and more diverse collections of energy assets, evaluate cross-domain transfer under limited adaptation data, and explore integration with federated optimization strategies. Further analysis of the learned latent space could also provide insight into the temporal structures and physical relationships captured by the representations. More broadly, these findings suggest that predictive self-supervised learning offers a promising direction for developing transferable and privacy-aware foundation models for energy time-series.

## 4. Conclusions

In this paper we presented a distributed JEPA framework for self-supervised representation learning from energy time-series. Our solution decouples representation learning from downstream forecasting tasks by predicting latent representations rather than reconstructing observations, while enabling integration of temporal and contextual information within a shared embedding space. Experiments on building energy consumption and photovoltaic generation datasets showed that the learned representations remained stable throughout training and avoided representation collapse. The resulting embeddings supported accurate 24-hour-ahead forecasting, achieving performance comparable to or exceeding a task-specific Transformer baseline. Also, our solution demonstrated strong generalization across photovoltaic sites and improved robustness under missing-data corruption. Overall, the results indicate that predictive latent-space objectives can learn transferable temporal representations without requiring task-specific supervision. However, our work was limited to building energy consumption and photovoltaic generation datasets and evaluated transferability through downstream forecasting tasks. Future work should investigate larger and more diverse energy domains, federated optimization strategies, and adaptation of the learned representations to additional tasks such as anomaly detection, or asset health monitoring.

### Acknowledgements

This work was supported by the project "Romanian Hub for Artificial Intelligence-HRIA", Smart Growth, Digitization and Financial Instruments Program, MySMIS, Romania no. 334906.